\documentclass{article}
\usepackage{spconf,amsmath,amssymb,graphicx}
\usepackage{booktabs}
\usepackage{multirow}
\usepackage{siunitx}
\usepackage[hidelinks]{hyperref}
\usepackage{microtype}
\usepackage{amsmath, amssymb}
\usepackage[ruled,vlined,linesnumbered]{algorithm2e}

\title{STRUCTURAL COMPLEXITY OF BRAIN MRI REVEALS AGE-ASSOCIATED PATTERNS}

\name{Anzhe Cheng$^{1 \dagger}$ \quad
Italo Ivo Lima Dias Pinto$^{2 \dagger}$ \quad 
Paul Bogdan$^1$
\thanks{$^\dagger$ These authors contributed equally to this work}}
\address{$^1$ University of Southern California, Los Angeles, CA, USA \\
$^2$ Instituto de Matemática e Estatística, Universidade de São Paulo, São Paulo, Brazil.
         }

\begin{document}
\maketitle

\begin{abstract}
We adapt structural complexity analysis to three-dimensional signals, with an emphasis on brain magnetic resonance imaging (MRI). This framework captures the multiscale organization of volumetric data by coarse-graining the signal at progressively larger spatial scales and quantifying the information lost between successive resolutions. While the traditional block-based approach can become unstable at coarse resolutions due to limited sampling, we introduce a sliding-window coarse-graining scheme that provides smoother estimates and improved robustness at large scales. Using this refined method, we analyze large structural MRI datasets spanning mid- to late adulthood and find that structural complexity decreases systematically with age, with the strongest effects emerging at coarser scales. These findings highlight structural complexity as a reliable signal processing tool for multiscale analysis of 3D imaging data, while also demonstrating its utility in predicting biological age from brain MRI.
\end{abstract}

\begin{keywords}
Structural complexity, brain MRI, aging, biomedical imaging
\end{keywords}

\section{Introduction}

Aging profoundly affects brain morphology, resulting in measurable changes in cortical thickness, subcortical volume, and large-scale structural organization that are visible on magnetic resonance imaging (MRI). These macroscopic biomarkers have been widely employed in studies of both typical and pathological aging, with large-scale initiatives such as ADNI, NACC, and UK Biobank (UKBB) assembling invaluable longitudinal datasets for this purpose \cite{petersen2010alzheimer,beekly2007national}. More recently, deep learning has enabled the development of ``brain-age'' models, which compare individual MRI scans to population-derived norms to quantify biological age. Recent efforts aim to produce Deep Neural Networks anatomically interpretable models that predict brain age while identifying domain-specific cognitive impairment \cite{yin2023anatomically}, quantify the pace of brain aging across cognitive domains \cite{yin2025deep}, and demonstrate accelerated senescence in older adults with traumatic brain injury \cite{Amgalan2022BrainAge}. Further advances explored saliency approaches for interpretability in both typical aging and traumatic brain injury, highlighting the neuroanatomical substrates most relevant for CNN predictions \cite{guo2024anatomic}. While these studies show that 3D convolutional neural networks (3D-CNNs) are powerful predictors, their reliance on saliency maps underscores an important gap: interpretability remains unstable across salience methods, with feature attribution varying depending on method. This limits our ability to extract principled, scale-aware biomarkers of aging from MRI.

Complexity science provides tools to quantify structural patterns distributed across multiple scales \cite{kantelhardt2002multifractal, peng1995quantification}. Applied to MRI, measures of spatial complexity can reveal subtle age-related or pathological alterations that may remain undetected by conventional volumetric metrics. Beyond their diagnostic potential, structural complexity measures also help address the interpretability challenges of CNNs: by identifying the spatial scales at which meaningful differences arise, they can guide the design of neural network architectures and highlight where multiscale feature extraction is most biologically informative.

In this paper, we propose the method of \emph{Multiscale structural complexity of natural patterns} \cite{bagrov2020multiscale} to three-dimensional brain MRI, refining the approach to improve stability at larger scales. Applying this measure to lifespan structural MRI, we find that age-related changes are consistently reflected in large-scale structural complexity: older individuals exhibit decreased 3D complexity compared to younger adults. This decline provides a novel biomarker that complements conventional volumetric and cortical thickness measures. In this way, structural complexity connects the explanatory richness of signal processing with the predictive power of neural networks, offering a path forward for more interpretable and biologically grounded models of brain aging.

\section{Methods}

\subsection{Structural Complexity calculation}

To quantify the multiscale organization of 3D scalar fields, we employed a structural complexity framework based on successive coarse-graining and overlap computation. The method operates by comparing the field $A$ with progressively coarser representations obtained at increasing spatial scales $\lambda_i = \lambda_0^{\,i+1}$, where $\lambda_0$ is the base scale and $i=1,\ldots,S$ indexes the coarse-graining step. Coarse-graining is applied in a cascading manner, such that the field at step $i$ is generated from the coarse-grained field of step $i-1$.  

For any field $X$ defined on a voxel lattice $\Omega$ of size $|\Omega|$, we denote spatial averages by
\begin{equation*}
    \langle X \rangle \;=\; \frac{1}{|\Omega|} \sum_{\mathbf{r}\in\Omega} X(\mathbf{r}).
\end{equation*}

With this convention, the overlap between two successive scales is defined as
\begin{equation*}
O(A_{\lambda_{i-1}}, B_{\lambda_i}) \;=\; \langle A_{\lambda_{i-1}} \, B_{\lambda_i} \rangle \;-\; \tfrac{1}{2}\Big(\langle A_{\lambda_{i-1}}^2 \rangle + \langle B_{\lambda_i}^2 \rangle \Big),
\end{equation*}
where $A_{\lambda_{i-1}}$ is the field at the previous step (with $A_{\lambda_0} = A$) and $B_{\lambda_i}$ its coarse-grained version at scale $\lambda_i$. Here, $\langle A_{\lambda_{i-1}} , B_{\lambda_i} \rangle$ denotes the spatial average of the voxelwise product between the fields $A_{\lambda_{i-1}}$ and $B_{\lambda_i}$. Note that $O(A,B) = -\tfrac{1}{2}\langle (A-B)^2 \rangle$, so the overlap is simply the negative mean squared deviation between successive resolutions. Its magnitude $|O|$ directly quantifies the information lost during coarse-graining: larger values indicate richer structural content at that scale. By construction, $O \leq 0$, and its magnitude quantifies the information lost when moving from one resolution level to the next.   

We define the \emph{scale-specific structural complexity} as
\begin{equation*}
C(\lambda_i) \;=\; \bigl| O(A_{\lambda_{i-1}}, B_{\lambda_i}) \bigr|.
\end{equation*}  

In the original formulation \cite{bagrov2020multiscale}, coarse-graining was performed through block averaging: the 3D volume is partitioned into non-overlapping cubic blocks of side $\lambda_i$, and each block is replaced by its mean value.

\begin{figure*}[!ht]
    \centering
    \includegraphics[width=\textwidth]{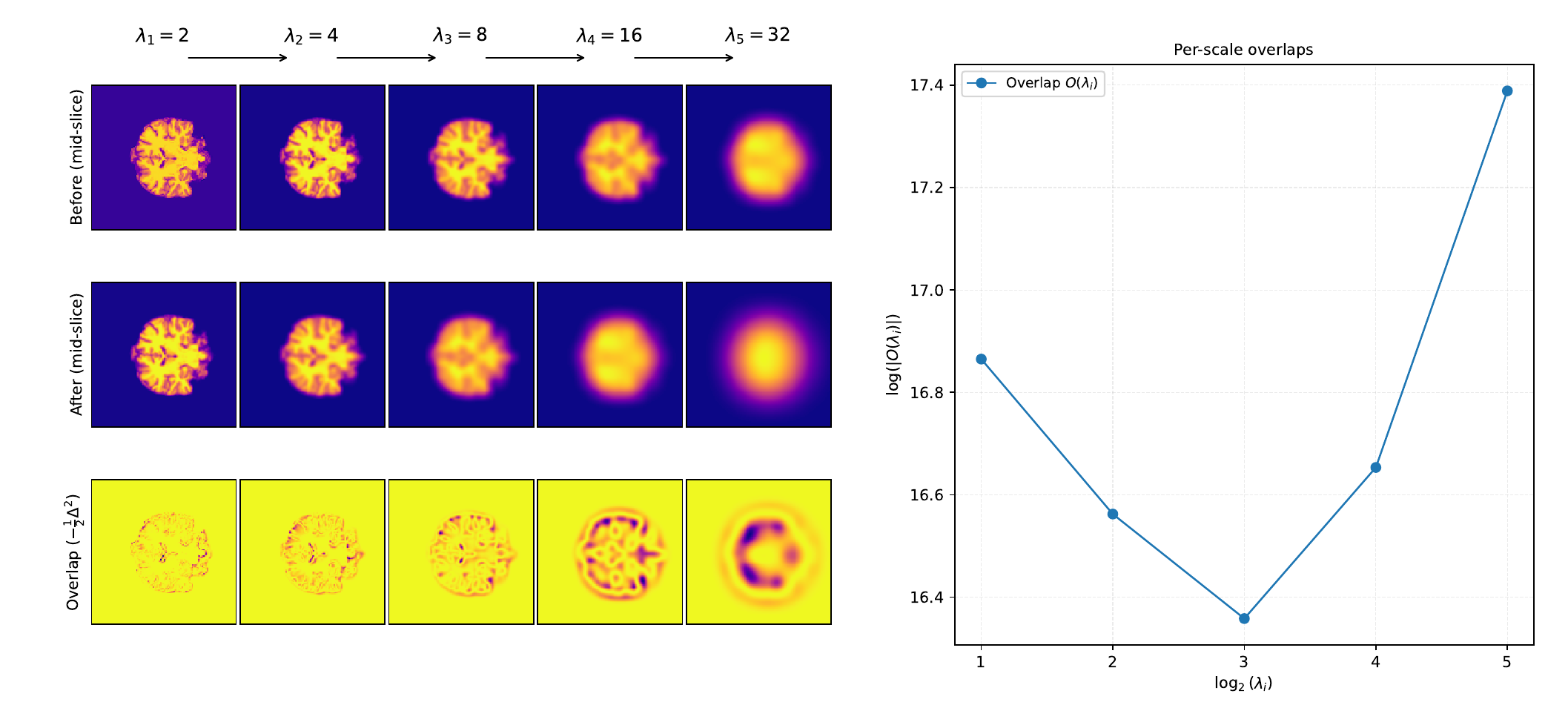}
    \caption{ Illustration of the structural complexity method for scalar fields. 
    Left panels show the mid-slice of the original volume (``Before''), its progressively coarse-grained representations at scales $\lambda_{1}=2$, $\lambda_{2}=4$, $\lambda_{3}=8$, $\lambda_{4}=16$, and $\lambda_{5}=32$ (``After''), and the corresponding pixel-wise overlaps between successive scales. 
    The right panel shows the per-scale overlap values $O(\lambda_i)$ as a function of the scale $\lambda_i$, displayed on logarithmic axes. 
    The cumulative contribution of these overlaps across scales defines the structural complexity measure.}
    \label{fig:sc_schematic}
\end{figure*}

\begin{algorithm}[!ht]
\DontPrintSemicolon
\label{SC_algorithm}
\caption{Multiscale Structural Complexity of a 3-D object}
\KwIn{
$V \in \mathbb{R}^{X \times Y \times Z}$; scales $\Lambda=\{\lambda_1<\cdots<\lambda_L\}$; \\
window $W=(w_x,w_y,w_z)$; stride $S=(s_x,s_y,s_z)$}
\KwOut{Complexity maps $\{K_{\lambda_k}\}_{k=1}^L$ and scale curve $\{C(\lambda_k)\}_{k=1}^L$}
\For{$k \leftarrow 1$ \KwTo $L$}{
  Pad $V$ as needed so each dimension is divisible by $\lambda_k$\;
  Partition into $\lambda_k \!\times\! \lambda_k \!\times\! \lambda_k$ blocks; replace each block by the \textbf{mean} of its voxels\;
  Reshape block means into downsampled volume $U \gets V_{\lambda_k}$ of size
  $X_k=\left\lfloor X/\lambda_k \right\rfloor$, $Y_k=\left\lfloor Y/\lambda_k \right\rfloor$, $Z_k=\left\lfloor Z/\lambda_k \right\rfloor$\;

  $N_x \leftarrow \left\lfloor (X_k - w_x)/s_x \right\rfloor + 1$,
  $N_y \leftarrow \left\lfloor (Y_k - w_y)/s_y \right\rfloor + 1$,
  $N_z \leftarrow \left\lfloor (Z_k - w_z)/s_z \right\rfloor + 1$\;
  Initialize $K_{\lambda_k} \in \mathbb{R}^{N_x \times N_y \times N_z}$\;

  \For{$i \leftarrow 1$ \KwTo $N_x$}{
    $\text{ox} \leftarrow (i-1)s_x$\;
    \For{$j \leftarrow 1$ \KwTo $N_y$}{
      $\text{oy} \leftarrow (j-1)s_y$\;
      \For{$\ell \leftarrow 1$ \KwTo $N_z$}{
        $\text{oz} \leftarrow (\ell-1)s_z$\;

        $B \leftarrow U[\text{ox}:(\text{ox}+w_x-1),\ \text{oy}:(\text{oy}+w_y-1),\ \text{oz}:(\text{oz}+w_z-1)]$\;

        $B_x \leftarrow B[2{:}w_x,\ 1{:}w_y,\ 1{:}w_z]$; \quad
        $B_y \leftarrow B[1{:}w_x,\ 2{:}w_y,\ 1{:}w_z]$; \quad
        $B_z \leftarrow B[1{:}w_x,\ 1{:}w_y,\ 2{:}w_z]$\;
        $B^\star \leftarrow B[1{:}(w_x\!-\!1),\ 1{:}(w_y\!-\!1),\ 1{:}(w_z\!-\!1)]$

$o_{B} \leftarrow \langle B^\star \!\cdot\! B^\star \rangle$\;

$o_{x} \leftarrow \langle B^\star \!\cdot\! B_x \rangle - \tfrac12\!\left(o_{B} + \langle B_x \!\cdot\! B_x \rangle\right)$\;
$o_{y} \leftarrow \langle B^\star \!\cdot\! B_y \rangle - \tfrac12\!\left(o_{B} + \langle B_y \!\cdot\! B_y \rangle\right)$\;
$o_{z} \leftarrow \langle B^\star \!\cdot\! B_z \rangle - \tfrac12\!\left(o_{B} + \langle B_z \!\cdot\! B_z \rangle\right)$\;
        $K_{\lambda_k}[i,j,\ell] \leftarrow -\tfrac{1}{3}\,(o_x + o_y + o_z)$\;
      }
    }
  }
  $C(\lambda_k) \leftarrow \mathrm{mean}\!\big(K_{\lambda_k}\big)$\;
}
\Return{$\{K_{\lambda_k}\}_{k=1}^L,\ \{C(\lambda_k)\}_{k=1}^L$}
\end{algorithm}

As a novel contribution, we introduced a sliding-window coarse-graining scheme. Here, each voxel is replaced by the mean over a cubic neighborhood of side $\lambda_i$ centered on that voxel. Because neighborhoods overlap, this scheme produces a smoothed, continuous representation without the tiling artifacts inherent to block averaging. Importantly, the sliding procedure improves the \textit{statistical stability} of the overlap estimates at large scales, where the block-based method suffers from having very few tiles to average over. This refinement ensures that the structural complexity remains reliable even as $\lambda_i$ approaches the size of the object. Algorithm \ref{SC_algorithm} presents a compact implementation of the method.

Figure \ref{fig:sc_schematic} illustrates the structural complexity procedure using the sliding-window scheme. Although the method is applied to the entire 3D brain structure, the left panels show only the mid-slice for illustration purposes. The three rows of plots on the left display the original scalar field together with its progressively coarse-grained representations at scales $\lambda_{1}=2$, $\lambda_{2}=4$, $\lambda_{3}=8$, $\lambda_{4}=16$, and $\lambda_{5}=32$. For each scale, we also show the voxel-wise overlap between the original and coarse-grained fields, highlighting the structural information captured across scales. The right panel presents the per-scale overlaps $O(\lambda_i)$ as a function of the coarse-graining scale $\lambda_i$ for this subject, revealing a scale-dependent pattern. As will be shown in the Results section, the largest scales ($\lambda_{4}=16$ and $\lambda_{5}=32$) carry the most relevant information for predicting brain age. This approach provides a multiscale measure of the structural richness of the object.

\subsection{Data and Preprocessing}

We analyzed structural T1-weighted brain MRI scans from three large cohorts: the UK Biobank (UKBB), the Alzheimer’s Disease Neuroimaging Initiative (ADNI)~\cite{petersen2010alzheimer}, and the National Alzheimer’s Coordinating Center (NACC) dataset~\cite{beekly2007national}. This combined dataset comprises on the order of thousands of MRI volumes, covering a broad age range from mid-adulthood through old age (approximately 44 to 90 years). The cohorts also include participants with cognitive impairment (351 with mild cognitive impairment and 359 with Alzheimer’s disease) to evaluate disease-specific effects. All scans were high-resolution 3D T1-weighted images acquired on 3.0 Tesla MRI scanners in UKBB, and on a mix of 1.5T and 3T systems in the multi-site ADNI and NACC studies. The native voxel resolution is approximately $1\times1\times1~\text{mm}^3$ for all datasets, providing detailed volumetric brain data

We followed the same preprocessing technique as Yin et al.~\cite{yin2025deep}. The preprocessing steps were performed consistently for all three cohorts to harmonize the data. Notably, NACC MRI data were collected from multiple sites with varying acquisition protocols, so applying the same preprocessing pipeline and intensity standardization across UKBB, ADNI, and NACC helped reduce inter-site variability. After preprocessing, the final brain volumes contained $128 \times 128 \times 128$ voxels and were saved as 3D NumPy arrays with shape $(128, 128, 128)$. These preprocessed volumes serve as the input for subsequent structural complexity analyses, with all images being in a common space and scale, and hence directly comparable across participants and cohorts.
\vspace{-5mm}
\section{Results}
\vspace{-3mm}
\begin{figure*}[!ht]
    \centering
    \includegraphics[width=\textwidth]{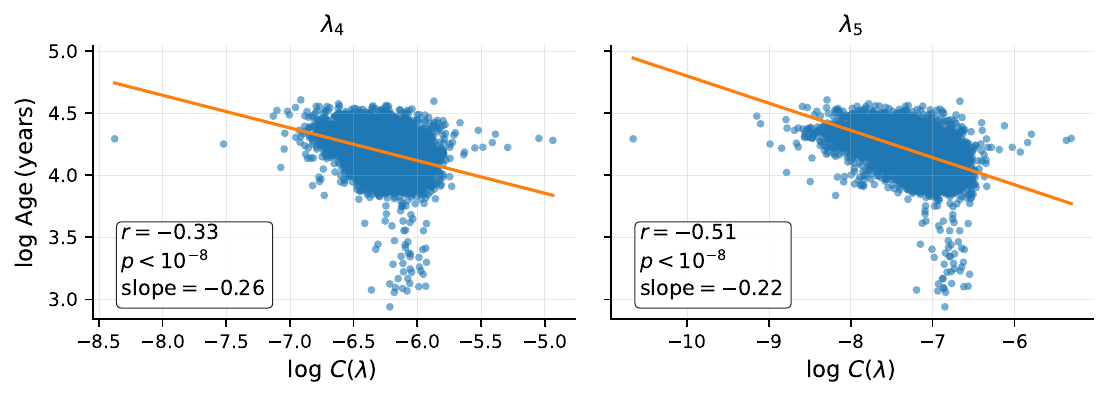}
    \caption{
        \textbf{Association between Age and Structural Complexity at Scales 4 and 5.} 
        Scatter plots show the relationship between $\log C(\lambda)$, the structural complexity of the brain signal at scale $\lambda$, and $\log$ Age (years). 
        Linear fits (orange lines) were estimated using least-squares regression in log--log space. 
        Pearson correlation coefficients ($r$) and FDR-corrected $p$-values are reported within each panel, indicating significant negative associations between age and complexity at both scales ($p < 0.001$ after correction). 
        These results suggest that structural complexity decreases systematically with aging at larger scales.
    }
    \label{fig:age_complexity}
\end{figure*}

To evaluate the relationship between structural complexity and aging, we applied our multiscale complexity analysis to the full dataset of participants’ brain signals. For each subject, structural complexity $C(\lambda)$ was computed across multiple scales $\lambda$, following the coarse-graining procedure described in the Methods. This yielded a scale-dependent estimate of signal organization, where lower values of $C(\lambda)$ reflect reduced structural heterogeneity and thus diminished multiscale richness.  

\begin{table*} [ht]
\centering
\caption{Correlation between age and structural complexity across scales $\lambda_0$ to $\lambda_5$.}
\begin{tabular}{c c c c c c c}
\hline
$\lambda$ & $r$ & $p$ & $q_{\mathrm{FDR}}$ & slope & intercept \\
\hline
$\lambda_{0}$ & -0.112 & $<10^{-8}$ & $<10^{-8}$ & -0.148 & 3.544 \\
$\lambda_{1}$ &  0.023 & $8.12\times 10^{-3}$  & $8.12\times 10^{-3}$  &  0.031 & 4.337 \\
$\lambda_{2}$ &  0.102 & $<10^{-8}$ & $<10^{-8}$ &  0.119 & 4.822 \\
$\lambda_{3}$ &  0.050 & $<10^{-8}$  & $<10^{-8}$  &  0.052 & 4.486 \\
$\lambda_{4}$ & -0.326 & $<10^{-8}$          & $<10^{-8}$          & -0.263 & 2.540 \\
$\lambda_{5}$ & -0.510 & $<10^{-8}$          & $<10^{-8}$          & -0.219 & 2.609 \\
\hline
\end{tabular}
\label{tab:results}
\end{table*}

The table \ref{tab:results} reports the correlation between age and structural complexity across scales 
$\lambda_0$ to $\lambda_5$. For each scale, the Pearson correlation coefficient ($r$), 
raw $p$-value, FDR-corrected $q$-value, and regression parameters (slope and intercept) 
are listed. Significant associations were observed at all scales after FDR correction 
($q<0.05$). However, at smaller scales ($\lambda_1$--$\lambda_3$), weak correlations were observed, 
whereas at larger scales ($\lambda_4$ and $\lambda_5$) stronger negative correlations 
emerge, indicating that structural complexity at coarser resolutions decreases with age.

Because both age and complexity exhibit multiplicative variability across the population, we analyzed the data in log-log space. Specifically, for each subject we calculated $\log C(\lambda)$ at scales $\lambda_{4}$ and $\lambda_{5}$ and compared these values against $\log$ Age. Linear regressions were fit to the scatter plots in order to quantify systematic dependencies, and Pearson’s correlation coefficients were computed to assess the strength of association. To control for multiple comparisons across scales, all reported $p$-values were adjusted using the Benjamini-Hochberg false discovery rate (FDR) correction.  

As shown in Figure~\ref{fig:age_complexity}, structural complexity decreased reliably with age at both scales. At $\lambda_{4}$, the relationship was characterized by a moderate negative correlation ($r=-0.33$, $p<10^{-8}$, slope $=-0.26$), such that increases in $\log C(\lambda)$ were associated with reductions in $\log$ Age. A stronger effect was observed at $\lambda_{5}$ ($r=-0.51$, $p<10^{-8}$, slope $=-0.22$), indicating that age-related reductions in complexity become more pronounced at larger scales of analysis. These findings can be directly understood in terms of the overlap definition: 
since $O(A,B) = -\tfrac{1}{2}\langle (A-B)^2 \rangle$, its magnitude $|O|$ measures the structural information lost when moving to a coarser representation. The observed decline of $|O|$ at the largest scales ($\lambda_4,\lambda_5$) in older adults therefore indicates that coarse-grained volumes become increasingly similar to the original images, revealing an age-related homogenization of large-scale brain structure.

\section{Conclusion}

We presented a refinement of the structural complexity framework, originally introduced in previous work \cite{bagrov2020multiscale}, by implementing a sliding-window coarse-graining scheme. This modification enhances the statistical stability of the method, particularly at the largest spatial scales $\lambda_i$, where the original block-based approach can suffer from limited sampling. By applying this approach to brain imaging data, we showed that structural complexity retains sensitivity to multiscale spatial organization while gaining robustness at coarse resolutions.

The practical implications for neuroscience and clinical imaging are significant. Structural complexity provides a complementary measure to traditional morphometric descriptors, capturing subtle multiscale features of cortical and subcortical organization that may serve as early indicators of aging and disease-related changes.

Finally, our results show how structural complexity can guide deep learning design. By identifying the scales $\lambda_i$ at which meaningful structural differences emerge, complexity analysis may guide the architecture of convolutional neural networks (CNNs). Suggesting biologically relevant filter sizes and multiscale feature extraction strategies. This provides a potential bridge between interpretable, theory-driven metrics and the performance of modern machine learning approaches.

\bibliographystyle{IEEEbib}
\bibliography{refs}

\end{document}